\documentclass[sigconf]{acmart}
\usepackage{subcaption}
\usepackage{booktabs}
\usepackage{multirow}
\usepackage{graphicx}
\usepackage{array}
\usepackage{caption}
\usepackage{pifont}
\AtBeginDocument{%
  }

\copyrightyear{2026}
\acmYear{2026}
\setcopyright{cc}
\setcctype{by}
\acmConference[MM '26]{Proceedings of the 34th ACM International Conference on Multimedia}{November 10--14, 2026}{Rio de Janeiro, Brazil}
\acmBooktitle{Proceedings of the 34th ACM International Conference on Multimedia (MM '26), November 10--14, 2026, Rio de Janeiro, Brazil}
\acmDOI{10.1145/3767308.3835300}
\acmISBN{979-8-4007-2213-4/2026/11}

\begin{document}

\title[What to Remove, What to Preserve: Dual-Ambiguity Rectification for All-in-One Image Restoration]{What to Remove, What to Preserve: Dual-Ambiguity Rectification for All-in-One Image Restoration}

\author{Cencen Liu}
\affiliation{%
  \institution{University of Electronic Science and Technology of China}
  \city{Chengdu}
  \state{Sichuan}
  \country{China}
}
\email{lcc@std.uestc.edu.cn}

\author{Wen Yin}
\affiliation{%
  \institution{University of Electronic Science and Technology of China}
  \city{Chengdu}
  \state{Sichuan}
  \country{China}
}
\email{yinwen1999@std.uestc.edu.cn}

\author{Dongyang Zhang}
\affiliation{%
  \institution{University of Electronic Science and Technology of China}
  \city{Chengdu}
  \state{Sichuan}
  \country{China}
}
\email{dyzhang@uestc.edu.cn}

\author{Dongmin Li}
\affiliation{%
  \institution{University of Electronic Science and Technology of China}
  \city{Chengdu}
  \state{Sichuan}
  \country{China}
}
\email{lidongmin@std.uestc.edu.cn}

\author{Shan Zhao}
\affiliation{%
  \institution{Jiigan Technology}
  \city{Chengdu}
  \state{Sichuan}
  \country{China}
}
\email{fly.shanzhao@gmail.com}

\author{Bing Su}
\affiliation{%
  \institution{Jiigan Technology}
  \city{Chengdu}
  \state{Sichuan}
  \country{China}
}
\email{subing@jiigan.com}

\author{Tao He}
\affiliation{%
  \institution{University of Electronic Science and Technology of China}
  \city{Chengdu}
  \state{Sichuan}
  \country{China}
}
\email{tao.he01@hotmail.com}

\author{Jielei Wang}
\affiliation{%
  \institution{University of Electronic Science and Technology of China}
  \city{Chengdu}
  \state{Sichuan}
  \country{China}
}
\email{jieleiwang\_uestc@163.com}

\author{Guoming Lu}
\authornote{Corresponding author.}
\affiliation{%
  \institution{University of Electronic Science and Technology of China}
  \city{Chengdu}
  \state{Sichuan}
  \country{China}
}
\email{lugm@uestc.edu.cn}

\renewcommand{\shortauthors}{Liu et al.}

\begin{abstract}
All-in-one image restoration aims to handle diverse degradations within a unified framework. Existing methods commonly encode heterogeneous degradation conditions in a shared latent space, where degradation-related cues and scene content can remain entangled. We characterize the resulting challenge as dual ambiguity: semantic ambiguity in channel-wise modulation and spatial ambiguity in restoration responses, which can lead to content corruption and residual artifacts. To mitigate this issue, we propose DAR-Net, a Dual-Ambiguity Rectification Network for all-in-one image restoration. DAR-Net first introduces a Degradation Archetype Representation (DAR) module to construct a structured degradation state through simplex-constrained archetype mixture modeling. Based on this state, a Semantic Ambiguity Rectification (SeAR) module generates degradation-aware prompts to improve channel-wise conditioning in the decoder. A Spatial Ambiguity Rectification (SpAR) module further regularizes degradation-aware and complementary features toward orthogonal response subspaces, reducing spatial interference between removal and preservation cues. Extensive experiments on standard all-in-one restoration benchmarks show that DAR-Net achieves the best overall performance under both three-degradation and five-degradation settings, improving the average PSNR over the strongest competitor by 0.14 dB and 0.34 dB, respectively; it additionally shows superior performance on CDD-11 and WeatherBench.
\end{abstract}

\ccsdesc[500]{Computing methodologies~Reconstruction}

\keywords{All-in-one image restoration, prompt-based restoration, degradation modeling, dual-ambiguity rectification}
\maketitle

\section{Introduction}
Image restoration aims to recover clean visual content from degraded observations affected by factors such as noise, haze, and rain conditions. As a front-end step in real-world multimedia content capture and enhancement pipelines, image restoration is also closely related to multimedia content quality. Most existing restoration methods are developed under a predefined degradation setting, and many representative models \cite{Chen2020PreTrainedIPT,NAFNet_eccv2022_chen, Restormer_cvpr2022_Zamir} are still deployed in a task-specific manner in practice. Such a paradigm leads to considerable computational and storage overhead and limits generalization in complex environments. To address this limitation, all-in-one image restoration (AIR) has emerged as a unified framework for handling diverse degradations within a single model \cite{AirNet_cvpr2022_Li, promptir_nips2023_Vaishnav, Instructir_eccv2024_Marcos}.

\begin{figure}[t]
    \captionsetup{aboveskip=0pt, belowskip=0pt}
    \centering
    \begin{subfigure}[b]{\linewidth}
        \centering
        \includegraphics[width=\linewidth]{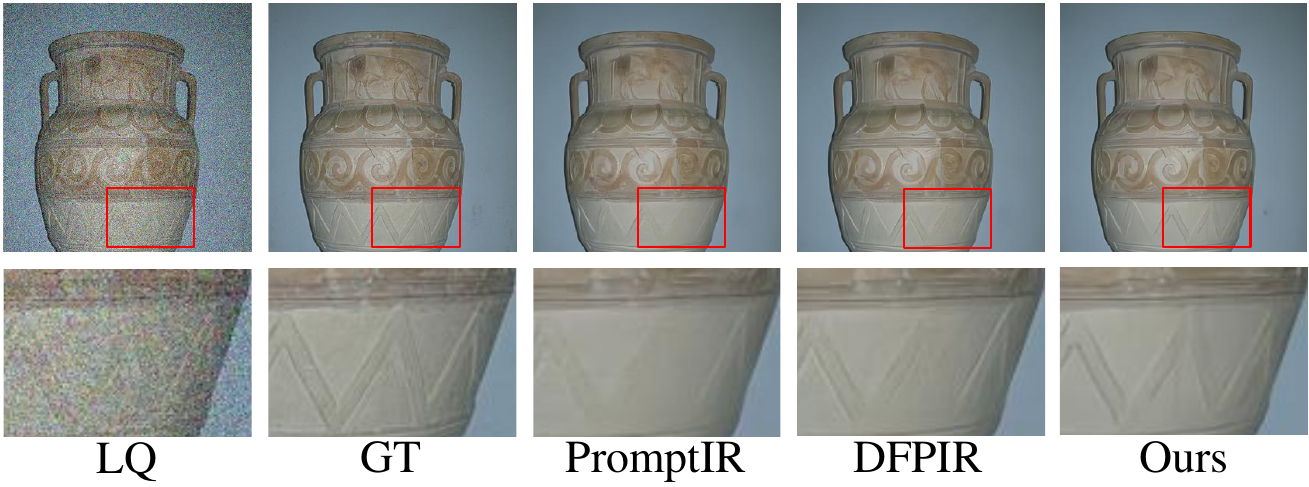}
        \caption{Content Corruption.}
        \label{fig:content_corruption}
    \end{subfigure}
    \begin{subfigure}[b]{\linewidth}
        \centering
        \includegraphics[width=\linewidth]{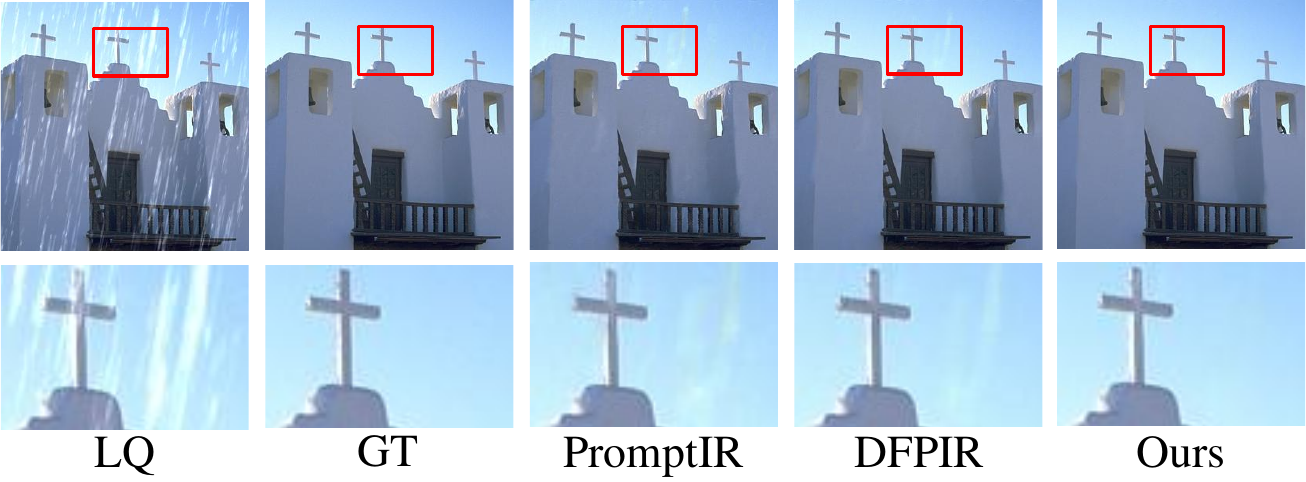}
        \caption{Residual Degradation.}
        \label{fig:residual_degradation}
    \end{subfigure}
    \caption{Typical failure modes caused by degradation-content entanglement. (a) \textit{Content Corruption}, where faithful image details are mistakenly suppressed. (b) \textit{Residual Degradation}, where degradation artifacts are retained.}
    \Description{Two example panels illustrate failure modes caused by degradation-content entanglement. The first shows content corruption, where faithful image details are suppressed, and the second shows residual degradation, where degradation artifacts remain.}
    \label{fig:failure_modes}
\end{figure}

Early AIR methods often rely on multi-branch designs, where different degradation types are handled by separate branches or task-specific modules. Although such designs improve degradation specialization, their parameter cost typically scales with the number of degradation types, which limits scalability and makes model deployment increasingly inefficient as restoration scenarios become more diverse \cite{Li2020AllIO,Han2021BlindID}. To improve scalability, subsequent studies increasingly adopt shared-backbone conditional restoration, where a unified network is modulated by degradation cues \cite{AirNet_cvpr2022_Li, promptir_nips2023_Vaishnav, DFPIR_CVPR2025_Tian}. Within this paradigm, different conditioning mechanisms have been explored. Prompt-based methods provide a representative and efficient solution by deriving lightweight conditioning signals from the input and injecting them into a shared backbone, thereby enabling input-adaptive restoration with limited additional parameters \cite{ai2024MPerceiver, jiang2024autodir, liu2025Uprestorer}. In parallel, other alternatives such as MoE-based routing enhance model capacity through dynamic expert selection, allowing different restoration patterns to be handled by different experts when necessary \cite{Yu2024MultiExpertAS, MoCEir_cvpr2024_Zamfir, AAAI2024MoFME}. However, although these methods differ in how they introduce degradation-aware conditioning, they still predominantly rely on shared latent representations within a unified restoration pipeline. As a result, a common difficulty remains insufficiently addressed: degradation-related cues and content-related representations are often encoded in an entangled manner within the shared latent space.

Therefore, existing unified restoration models still face difficulty in distinguishing what should be removed from what should be preserved. This challenge is often reflected in two typical failure modes, as shown in Fig.~\ref{fig:failure_modes}: (1) \textit{Content Corruption}, where faithful image content is mistakenly removed together with degradation, and (2) \textit{Residual Degradation}, where degradation patterns are not sufficiently suppressed in the restored image. We analyze these failures through two forms of ambiguity. The first is \textit{Semantic Ambiguity}, namely the channel-wise entanglement between degradation-related cues and content-related representation, which can make degradation-aware modulation less discriminative. The second is \textit{Spatial Ambiguity}, namely their entanglement in the spatial dimension, which can weaken the spatial selectivity of restoration responses.

Motivated by these observations, we propose a Dual-Ambiguity Rectification Network \textbf{(DAR-Net)} for all-in-one image restoration. DAR-Net reduces degradation-content entanglement in both the channel and spatial dimensions through three components. It first employs a \textbf{Degradation Archetype Representation (DAR)} module to construct an archetype-based degradation state, which serves as a degradation prior for subsequent rectification. Based on this degradation state, the \textbf{Semantic Ambiguity Rectification (SeAR)} module alleviates channel-wise entanglement through an Archetype-Guided Prompt Generator (AGPG) and a Degradation-Aware Prompt Integrator (DAPI). Specifically, AGPG first generates a base prompt and then refines it via archetype-guided channel routing conditioned on the degradation state, yielding a degradation-aware prompt. DAPI subsequently injects this rectified prompt into the restoration process for degradation-conditioned feature modulation. In addition, the \textbf{Spatial Ambiguity Rectification (SpAR)} module reduces spatial entanglement via \textbf{Orthogonal Subspace Rectification (OSR)}, which treats the rectified prompt as a degradation-aware representation and derives a complementary content-related feature from the latent representation, encouraging the two to occupy orthogonal subspaces. Together, these designs aim to better distinguish removal-related and preservation-related cues in image restoration.

The main contributions are summarized as follows:
\begin{itemize}
    \item \textbf{Dual-ambiguity rectification framework:} We propose DAR-Net for AIR, which mitigates degradation-content entanglement in both the channel and spatial dimensions.

    \item \textbf{Semantic ambiguity rectification:} We introduce DAR to model degradation states and SeAR to generate degradation-aware prompts for channel-wise semantic rectification.

    \item \textbf{Spatial ambiguity rectification:} We introduce SpAR with OSR to separate degradation-related and complementary content-related representations in the spatial dimension.

    \item \textbf{Comprehensive validation and results:} Extensive experiments verify that DAR-Net consistently achieves the best overall performance on standard all-in-one restoration benchmarks and generalizes favorably to mixed and real-world degradations.
\end{itemize}

\begin{figure*}[!t]
    \centering
    \includegraphics[width=1\linewidth]{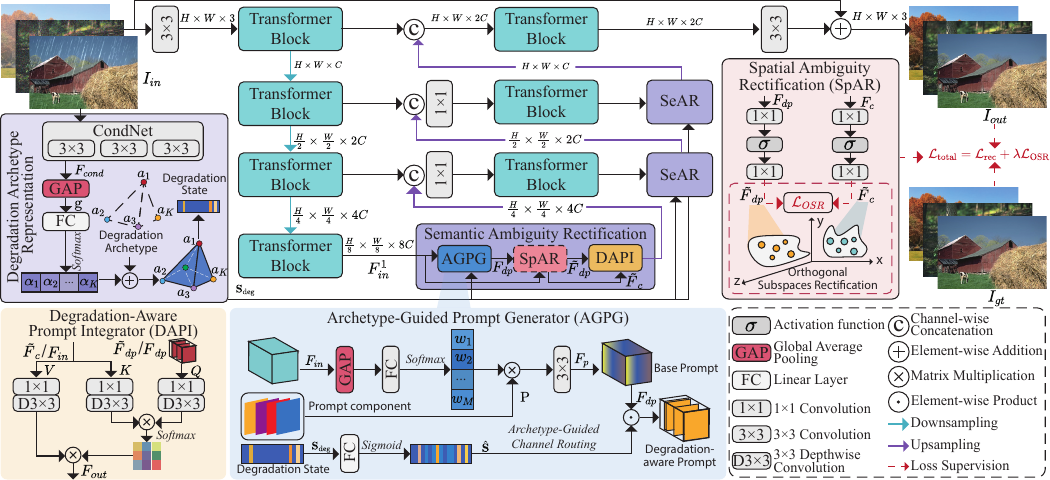}
    \caption{Architecture of the Dual-Ambiguity Rectification Network (DAR-Net). Built upon a hierarchical U-shaped Transformer, DAR-Net mitigates degradation-content entanglement through three components: (1) \textbf{DAR}, which constructs a degradation state via simplex-constrained degradation archetype representation; (2) \textbf{SeAR}, which comprises AGPG to generate a degradation-aware prompt from the degradation state and DAPI to integrate this prompt into the decoder for degradation-conditioned restoration; and (3) \textbf{SpAR}, which is implemented by OSR to regularize the degradation-aware prompt and the corresponding content feature toward orthogonal spatial subspaces under the $\mathcal{L}_{\mathrm{OSR}}$ constraint.}
    \Description{Block diagram of DAR-Net. A hierarchical U-shaped Transformer processes the input image. DAR constructs a simplex-constrained degradation state, SeAR uses AGPG and DAPI to generate and integrate degradation-aware prompts in the decoder, and SpAR uses OSR to encourage the prompt and content feature to occupy orthogonal spatial subspaces.}
    \vspace{-5pt}
    \label{fig:arch}
\end{figure*}

\section{Related Work}

\subsection{Task-Specific Image Restoration}
Traditional image restoration methods are typically designed for a single degradation type, such as denoising~\cite{Liang2021SwinIRIR,NAFNet_eccv2022_chen,Shen2022AdaptiveDF}, deblurring~\cite{Kong2022EfficientFD,Whang2021DeblurringVS,Cho2021RethinkingCA}, deraining~\cite{Chen2023LearningAS,Wang2020JointSA,Yi2021StructurePreservingDW,Xiao2022ImageDT}, and dehazing~\cite{Chen2023DEANetSI,Shen2023MutualIT,Song2022VisionTF,Cai2016DehazeNetAE}. Their goal is to learn a direct mapping from degraded images to clean images under a predefined degradation setting. With the development of restoration architectures, many methods have gradually moved from heavily customized task-specific designs toward more general restoration backbones. Representative models such as IPT~\cite{Chen2020PreTrainedIPT}, SwinIR~\cite{Liang2021SwinIRIR}, Uformer~\cite{Wang2021UformerAG}, Restormer~\cite{Restormer_cvpr2022_Zamir}, NAFNet~\cite{NAFNet_eccv2022_chen}, and MAXIM~\cite{Tu2022MAXIMMM} exemplify this trend by improving restoration quality through stronger feature modeling and broader contextual interaction. Recent image super-resolution methods further explore hybrid Mamba--Transformer modeling to improve efficient long-range interaction~\cite{Liu2025SRMambaT}. While task-specific image restoration methods are effective for individual degradation types, their reliance on separate degradation-specific models leads to poor scalability and limits their applicability in unified real-world restoration settings. 

\subsection{All-in-one Image Restoration}
\begingroup
\emergencystretch=1em
To improve scalability in unified real-world restoration scenarios, all-in-one image restoration aims to handle diverse degradations with a single unified model, making it more suitable for practical settings where the degradation type is unknown or mixed. Existing methods mainly differ in how degradation information is incorporated into a shared restoration pipeline. Representative early directions include degradation representation learning, where AirNet learns contrastive degradation representations~\cite{AirNet_cvpr2022_Li}; prompt-based conditioning, where PromptIR~\cite{promptir_nips2023_Vaishnav}, InstructIR~\cite{Instructir_eccv2024_Marcos}, and UP-Restorer~\cite{liu2025Uprestorer} inject learned prompts or instructions into the restoration network; and multimodal guidance, where DA-CLIP~\cite{DACLIP_ICLR2024_Luo} and MPerceiver~\cite{ai2024MPerceiver} leverage large-scale vision-language priors for restoration. Subsequent works further improve unified restoration either by strengthening degradation modeling and shared representation learning~\cite{Chen2025UniRestore,CVPR2024language,jiang2024autodir,Zhang2023IngredientorientedML,tpami2026BaryIR} or by introducing degradation-specialized experts to better handle diverse degradation patterns~\cite{MoCEir_cvpr2024_Zamfir,AAAI2024MoFME,Wang2025M2RestoreMM}. Despite these different designs, most existing methods still rely on degradation cues to condition, organize, or route shared features within a unified restoration network. In contrast, our method focuses on reducing degradation-content entanglement during feature modulation, so that the model can better distinguish what should be removed from what should be preserved.
\par
\endgroup

\section{Methods}
\subsection{Mathematical Preliminaries}
\label{sec:pre}

\noindent\textbf{Simplices and barycentric coordinates.}
Let $\mathcal{V}=\{v_1,\dots,v_K\}\subset\mathbb{R}^D$ with $K\le D+1$. If $v_1,\dots,v_K$ are affinely independent, then their convex hull
\begin{equation}
\operatorname{conv}(\mathcal{V})
=
\left\{
\sum_{k=1}^K \alpha_k v_k
\;\middle|\;
\alpha_k\ge 0,\;
\sum_{k=1}^K \alpha_k=1
\right\}
\end{equation}
forms a geometric $(K-1)$-simplex. Equivalently, letting
\begin{equation}
\Delta^{K-1}
=
\left\{
\alpha\in\mathbb{R}^K
\;\middle|\;
\alpha_k\ge 0,\;
\sum_{k=1}^K \alpha_k=1
\right\},
\end{equation}
the affine map
\begin{equation}
T:\Delta^{K-1}\to\operatorname{conv}(\mathcal{V}),
\qquad
T(\alpha)=\sum_{k=1}^K \alpha_k v_k
\end{equation}
is bijective. Hence every point $z\in\operatorname{conv}(\mathcal{V})$ admits a unique coefficient vector $\alpha\in\Delta^{K-1}$ such that
\begin{equation}
z=\sum_{k=1}^K \alpha_k v_k,
\end{equation}
where $\alpha$ is the barycentric coordinate of $z$ with respect to $\mathcal{V}$.

\noindent\textbf{Orthogonal decomposition.}
Let $\mathcal{H}$ be a finite-dimensional inner-product space with inner product $\langle\cdot,\cdot\rangle$. For any subspace $\mathcal{S}\subseteq\mathcal{H}$, its orthogonal complement is
\begin{equation}
\mathcal{S}^{\perp}
=
\left\{
y\in\mathcal{H}
\;\middle|\;
\langle x,y\rangle=0,\ \forall x\in\mathcal{S}
\right\}.
\end{equation}
By the projection theorem, every $z\in\mathcal{H}$ admits a unique orthogonal decomposition
\begin{equation}
z=P_{\mathcal{S}}z+P_{\mathcal{S}^{\perp}}z,
\qquad
P_{\mathcal{S}}z\in\mathcal{S},\;
P_{\mathcal{S}^{\perp}}z\in\mathcal{S}^{\perp},
\end{equation}
which implies
\begin{equation}
\|z\|_2^2
=
\|P_{\mathcal{S}}z\|_2^2
+
\|P_{\mathcal{S}^{\perp}}z\|_2^2.
\end{equation}
More generally, if $\mathcal{S}_A\perp\mathcal{S}_B$, then for any $x_A\in\mathcal{S}_A$ and $x_B\in\mathcal{S}_B$,
\begin{equation}
\langle x_A,x_B\rangle=0,
\qquad
\|x_A+x_B\|_2^2=\|x_A\|_2^2+\|x_B\|_2^2.
\end{equation}

In the Euclidean case, let $X_A,X_B\in\mathbb{R}^{C\times N}$, and denote their row vectors by $\{x_i\}_{i=1}^C$ and $\{y_j\}_{j=1}^C$. Then the $(i,j)$-th entry of the cross-Gram matrix satisfies
\begin{equation}
[X_A X_B^\top]_{ij}=\langle x_i,y_j\rangle.
\end{equation}
Therefore, the row-generated subspaces of $X_A$ and $X_B$ are orthogonal if and only if
\begin{equation}
X_A X_B^\top=\mathbf{0}_{C\times C}.
\end{equation}
Moreover,
\begin{equation}
\|X_A X_B^\top\|_F^2
=
\sum_{i=1}^C\sum_{j=1}^C \langle x_i,y_j\rangle^2,
\end{equation}
which measures the total pairwise interaction between the two subspaces and vanishes exactly under orthogonality.

\subsection{Overview}
Built upon a hierarchical U-shaped Transformer backbone, DAR-Net mitigates two ambiguities in AIR (Fig.~\ref{fig:arch}). Specifically, we design a rectification pipeline: (1) \textbf{Degradation Archetype Representation (DAR)} (\S~\ref{sec:dar}) extracts a global degradation descriptor and maps it to simplex-constrained mixture coefficients to construct a degradation state; (2) \textbf{Semantic Ambiguity Rectification (SeAR)} (\S~\ref{sec:sear}) uses this state to rectify prompt channels, yielding a degradation-aware prompt that is further integrated into the decoder for degradation-conditioned feature modulation; and (3) \textbf{Spatial Ambiguity Rectification (SpAR)} (\S~\ref{sec:spar}) regularizes the degradation-aware prompt and the corresponding content feature at the deepest decoder stage toward orthogonal subspaces. Finally, the restored image is reconstructed with a global residual connection, and the training objective is given in \S~\ref{sec:loss}.

\subsection{Degradation Archetype Representation}
\label{sec:dar}

The DAR module implements the simplex-constrained parameterization introduced in \S~\ref{sec:pre} and provides a structured degradation representation for subsequent rectification. Given an input image $I_{in}\in\mathbb{R}^{3\times H\times W}$, where $H$ and $W$ denote the input height and width, respectively, we first employ a lightweight conditioning network $\mathrm{Cond}(\cdot)$, implemented by stacked $3\times3$ convolutional layers, to extract degradation-sensitive features:
\begin{equation}
F_{cond}=\mathrm{Cond}(I_{in}), \qquad F_{cond}\in\mathbb{R}^{C\times H\times W},
\end{equation}
where $C$ is the channel dimension. We then use global average pooling to $F_{cond}$ to obtain a degradation descriptor $\mathbf{g}\in\mathbb{R}^{C}$. This descriptor is projected to a $K$-dimensional score vector and normalized by a softmax operator to produce the mixture coefficients
\begin{equation}
\boldsymbol{\alpha}=\mathrm{Softmax}(\mathbf{W}\mathbf{g}+\mathbf{b}),
\qquad
\boldsymbol{\alpha}\in\mathbb{R}^{K},
\end{equation}
where $\mathbf{W}\in\mathbb{R}^{K\times C}$, $\mathbf{b}\in\mathbb{R}^{K}$, and $K$ denotes the number of degradation archetypes. Here, each entry of $\boldsymbol{\alpha}$ quantifies the contribution of one archetype to the degradation mixture, and thus $\boldsymbol{\alpha}$ serves as the simplex coordinate vector described in \S~\ref{sec:pre}. Let $\mathbf{A}=[\mathbf{a}_1,\mathbf{a}_2,\dots,\mathbf{a}_K]\in\mathbb{R}^{C\times K}$ denote a learnable degradation archetype matrix, where each column $\mathbf{a}_k\in\mathbb{R}^{C}$ is an archetypal degradation vector. The degradation state is then constructed as
\begin{equation}
\mathbf{S}_{\mathrm{deg}}=\sum_{k=1}^{K}\alpha_k\mathbf{a}_k=\mathbf{A}\boldsymbol{\alpha},
\qquad
\mathbf{S}_{\mathrm{deg}}\in\mathbb{R}^{C},
\end{equation}
which is a simplex-constrained convex combination of the learned archetypes. As a result, $\mathbf{S}_{\mathrm{deg}}$ lies in the convex hull of the archetypes and serves as the structured degradation representation used in the subsequent rectification modules.

\subsection{Semantic Ambiguity Rectification}
\label{sec:sear}

The SeAR mitigates semantic ambiguity, i.e., the channel-wise entanglement between degradation representation and content representation. SeAR consists of an Archetype-Guided Prompt Generator (AGPG) and a Degradation-Aware Prompt Integrator (DAPI). Specifically, SeAR first uses the degradation state $\mathbf{S}_{\mathrm{deg}}$ to generate a degradation-aware prompt, and then integrates this prompt into the stage-wise decoding process. Before the deepest decoder stage, SeAR further derives a content feature by residual decomposition.

\subsubsection{Archetype-Guided Prompt Generator (AGPG)}

Formally, let $F_{in}^{(l)} \in \mathbb{R}^{C_l \times H_l \times W_l}$ denote the input feature before the $l$-th decoder stage ($l\in\{1,2,3\}$). AGPG aims to construct a degradation-aware prompt by combining two sources of information: the current-stage feature, which provides input-adaptive prompt cues, and the degradation state $\mathbf{S}_{\mathrm{deg}}$, which provides structured degradation prior. To this end, we first synthesize a base prompt from a set of $M$ learnable prompt tensors
$\mathcal{P}^{(l)}=\{P_1^{(l)},\dots,P_M^{(l)}\}$, where each
$P_m^{(l)} \in \mathbb{R}^{C_l \times H_l \times W_l}$. Specifically, we predict an input-dependent mixture weight vector from the globally pooled feature and use it to aggregate the prompt tensors:
\begin{equation}
\begin{aligned}
    \mathbf{w}^{(l)}
    &= \mathrm{Softmax}\!\left(\mathrm{FC}_{p}^{(l)}(\mathcal{G}(F_{in}^{(l)}))\right),\\
    F_{p}^{(l)}
    &= \mathrm{Conv}_{3\times3}^{(l)}\!\left(\sum_{m=1}^{M} w_m^{(l)} P_m^{(l)}\right).
\end{aligned}
\end{equation}

\begingroup
\emergencystretch=1em
We then inject the degradation prior by mapping $\mathbf{S}_{\mathrm{deg}}$ to a channel-wise routing vector and using it to rectify the base prompt:
\begin{equation}
    \hat{\mathbf{s}}^{(l)} = \sigma\!\left(\mathrm{FC}_{s}^{(l)}(\mathbf{S}_{\mathrm{deg}})\right), \qquad
    F_{dp}^{(l)} = F_{p}^{(l)} \odot \hat{\mathbf{s}}^{(l)},
\end{equation}
where $\hat{\mathbf{s}}^{(l)} \in \mathbb{R}^{C_l \times 1 \times 1}$ and $\odot$ denotes broadcast multiplication over spatial dimensions. In this way, channels that are more consistent with the inferred degradation state are emphasized, while prompt responses unrelated to the current degradation are suppressed.
\par
\endgroup

At the deepest decoder stage, we further derive a complementary content-related feature by residual subtraction,
\begin{equation}
    F_{c}^{(1)} = F_{in}^{(1)} - F_{dp}^{(1)},
\end{equation}
and forward the pair $(F_{dp}^{(1)}, F_{c}^{(1)})$ to SpAR for subsequent spatial ambiguity rectification.

\begin{table*}[t]\tiny
\centering
\caption{Quantitative comparison on the 3D all-in-one restoration benchmark. We reported PSNR/SSIM.}
\vspace{-10pt}
\resizebox{\textwidth}{!}{
\begin{tabular}{lccccccc}
\toprule
\multirow{2}{*}[-2.5pt]{\textbf{Method}} & \multirow{2}{*}[-2.5pt]{\textbf{Venue}}  & \textbf{Dehazing} & \textbf{Deraining} & \multicolumn{3}{c}{\textbf{Denoising on BSD68}} & \multirow{2}{*}[-2.5pt]{\textbf{Average}} \\
\cmidrule(lr){5-7}
& & SOTS-Outdoor & Rain100L & $\sigma=15$ & $\sigma=25$ & $\sigma=50$ & \\
\midrule
Restormer \cite{Restormer_cvpr2022_Zamir}& CVPR'22  & 27.78/0.958 & 33.78/0.958 & 33.72/0.930 & 30.67/0.865 & 27.63/0.792 & 30.75/0.901 \\
AirNet \cite{AirNet_cvpr2022_Li}& CVPR'22 & 27.94/0.962 & 34.90/0.968 & 33.92/0.933 & 31.26/0.888 & 28.00/0.797 & 31.20/0.910 \\
PromptIR \cite{promptir_nips2023_Vaishnav}& NeurIPS'23 & 30.58/0.974 & 36.37/0.972 & 33.98/0.933 & 31.31/0.888 & 28.06/0.799 & 32.06/0.913 \\
InstructIR \cite{Instructir_eccv2024_Marcos}& ECCV'24 & 30.22/0.959 & 37.98/0.978 & {34.15}/0.933 & {31.52}/0.890 & 28.30/0.804 & 32.43/0.913 \\
DiffUIR \cite{DiffUIR_CVPR2024_Zheng}& CVPR'24 & 30.18/0.973 & 36.78/0.973 & 33.94/0.932 & 31.26/0.887 & 28.04/0.797 & 32.04/0.912 \\
AdaIR \cite{AdaIR_ICLR2025_Cui}& ICLR'25 & 31.06/0.980 & 38.64/0.983 & 34.12/\underline{0.935} & 31.45/0.892 & 28.19/0.802 & 32.69/0.918 \\
VLU-Net \cite{VLUNet_cvpr2025_Zeng}& CVPR'25 & 30.71/0.980 & {38.93}/{0.984} & 31.13/\underline{0.935} & 31.48/0.892 & 28.23/0.804 & 32.10/{0.919} \\
MoCE-IR \cite{MoCEir_cvpr2024_Zamfir}& CVPR'25 & 31.34/0.979 & 38.57/{0.984} & 34.11/0.932 & 31.45/0.888 & 28.18/0.800 & 32.73/0.917 \\
DFPIR \cite{DFPIR_CVPR2025_Tian}& CVPR'25 & \underline{31.87}/0.980 & 38.65/0.982 & 34.12/\underline{0.935} & 31.47/\underline{0.893} & 28.25/\underline{0.806} & 32.88/{0.919} \\
ClearAIR \cite{AAAI2026ClearAIR}& AAAI'26 & 31.08/\underline{0.981} & 38.61/0.984 & \underline{34.18}/\underline{0.935} & \underline{31.50}/0.891 & \underline{28.31}/0.804 & 32.74/0.919 \\
 MIRAGE \cite{ICLR2026MIRAGE}& ICLR'26 & 31.86/\underline{0.981} & \underline{38.94}/\underline{0.985} & 34.12/\underline{0.935} & 31.46/0.891 & 28.19/0.803 & \underline{32.91}/0.919 \\
\midrule
{DAR-Net (Ours)} & - & \textbf{31.93}/\textbf{0.984} & \textbf{39.15}/\textbf{0.986} & \textbf{34.21}/\textbf{0.936} & \textbf{31.58}/\textbf{0.895} & \textbf{28.37}/\textbf{0.808} & \textbf{33.05}/\textbf{0.922} \\
\bottomrule
\end{tabular}
}
\label{tab:main_3d}
\end{table*}

\subsubsection{Degradation-Aware Prompt Integrator (DAPI)}

DAPI injects the degradation-aware prompt into the decoder through channel-wise attention. For a unified formulation, we define
\begin{equation}
F_{q}^{(l)}=
\begin{cases}
\tilde{F}_{dp}^{(1)}, & l=1,\\
F_{dp}^{(l)}, & l>1,
\end{cases}
\qquad
F_{kv}^{(l)}=
\begin{cases}
\tilde{F}_{c}^{(1)}, & l=1,\\
F_{in}^{(l)}, & l>1.
\end{cases}
\end{equation}
Here, the deepest decoder stage uses the SpAR-rectified pair $(\tilde{F}_{dp}^{(1)},\allowbreak \tilde{F}_{c}^{(1)})$, while later stages directly use the degradation-aware prompt and the current-stage input feature. We then project these inputs into query, key, and value tensors:
\begin{equation}
    Q^{(l)} = \Phi_Q^{(l)}(F_{q}^{(l)}), \qquad
    K^{(l)} = \Phi_K^{(l)}(F_{kv}^{(l)}), \qquad
    V^{(l)} = \Phi_V^{(l)}(F_{kv}^{(l)}),
\end{equation}
where $\Phi_Q^{(l)}$, $\Phi_K^{(l)}$, and $\Phi_V^{(l)}$ are three independent projection blocks, each implemented by a $1\times1$ convolution followed by a depth-wise $3\times3$ convolution. Let $N_l = H_l W_l$ denote the number of spatial locations at the $l$-th stage. After reshaping $Q^{(l)}$, $K^{(l)}$, and $V^{(l)}$ to $\mathbb{R}^{C_l \times N_l}$, we compute channel-wise attention as
\begin{equation}
    A^{(l)} = \mathrm{Softmax}\!\left(\frac{Q^{(l)} {K^{(l)}}^{\top}}{\tau}\right),
    \qquad A^{(l)} \in \mathbb{R}^{C_l \times C_l},
\end{equation}
where $\tau$ is a learnable temperature parameter and the softmax is applied row-wise. The stage output is then obtained as $F_{out}^{(l)} = \mathrm{Reshape}\!\left(A^{(l)} V^{(l)}\right).$

\subsection{Spatial Ambiguity Rectification}
\label{sec:spar}

SeAR produces a degradation-aware prompt $F_{dp}^{(1)}$ and a complementary  content-related feature $F_{c}^{(1)}$. Although residual decomposition separates them coarsely, their spatial responses may still remain entangled, leading to spatial ambiguity. To further separate degradation-related and content-related spatial responses before prompt integration, we introduce an \textbf{Orthogonal Subspace Rectification (OSR)} strategy, which encourages the two representations to lie in orthogonal subspaces. Specifically, we first transform $F_{dp}^{(1)}$ and $F_{c}^{(1)}$ with two learnable mappings $\Omega_p$ and $\Omega_c$ while preserving their spatial resolution:
\begin{equation}
    \tilde{F}_{dp}^{(1)} = \Omega_p(F_{dp}^{(1)}), \qquad
    \tilde{F}_{c}^{(1)} = \Omega_c(F_{c}^{(1)}),
\end{equation}
where $\tilde{F}_{dp}^{(1)}, \tilde{F}_{c}^{(1)} \in \mathbb{R}^{C_1 \times H_1 \times W_1}$. In practice, each mapping is implemented by a $1\times1$ convolution, a point-wise nonlinearity, and another $1\times1$ convolution. These learnable transformations allow the model to project the two features into a space where orthogonality can be imposed more effectively.

To instantiate the orthogonality constraint, let $N_1 = H_1 W_1$, and let $\tilde{\mathbf{m}}_{dp,k}, \tilde{\mathbf{m}}_{c,k}\in\mathbb{R}^{N_1}$ denote the flattened spatial maps of the $k$-th channel of $\tilde{F}_{dp}^{(1)}$ and $\tilde{F}_{c}^{(1)}$, respectively. Since directly shrinking feature magnitudes could trivially reduce their interaction, we first normalize each channel vector:
\begin{equation}
    \hat{\mathbf{m}}_{dp,k}
    =
    \frac{\tilde{\mathbf{m}}_{dp,k}}{\|\tilde{\mathbf{m}}_{dp,k}\|_2+\epsilon},
    \qquad
    \hat{\mathbf{m}}_{c,k}
    =
    \frac{\tilde{\mathbf{m}}_{c,k}}{\|\tilde{\mathbf{m}}_{c,k}\|_2+\epsilon},
\end{equation}
where $\epsilon$ is a constant for numerical stability. We then stack the normalized vectors row-wise into
$X_{dp}, X_c \in \mathbb{R}^{C_1 \times N_1}$. In this form, the row spaces of $X_{dp}$ and $X_c$ represent the spatial response subspaces of the degradation-aware and content features, respectively. According to the orthogonal decomposition in \S~\ref{sec:pre}, two row-generated subspaces are orthogonal if and only if their cross-Gram matrix vanishes, i.e., $X_{dp}X_c^\top=\mathbf{0}$. We therefore define the OSR loss as
\begin{equation}
    \mathcal{L}_{\mathrm{OSR}}
    =
    \|X_{dp}X_c^\top\|_F^2
    =
    \sum_{i=1}^{C_1}\sum_{j=1}^{C_1}
    \langle \hat{\mathbf{m}}_{dp,i}, \hat{\mathbf{m}}_{c,j}\rangle^2.
    \label{eq:OSR}
\end{equation}
Minimizing $\mathcal{L}_{\mathrm{OSR}}$ suppresses all pairwise inner-product interactions between the channel-wise spatial responses of the two features, thereby encouraging their row-generated subspaces to be orthogonal. The resulting rectified representations $\tilde{F}_{dp}^{(1)}$ and $\tilde{F}_{c}^{(1)}$ are then fed into DAPI at the deepest decoder stage.

\begin{table*}[t]\tiny
\centering
\caption{Quantitative comparison on the 5D all-in-one restoration benchmark. We reported PSNR/SSIM.}
\vspace{-10pt}
\resizebox{\textwidth}{!}{
\begin{tabular}{lccccccc}
\toprule
\multirow{2}{*}{\textbf{Method}} & \multirow{2}{*}{\textbf{Venue}} & \textbf{Dehazing} & \textbf{Deraining} & \textbf{Denoising} & \textbf{Deblurring} & \textbf{Low-Light} & \multirow{2}{*}{\textbf{Average}} \\
 & & SOTS-Outdoor & Rain100L & BSD68 ($\sigma=25$) & GoPro & LOL & \\
\midrule
Restormer \cite{Restormer_cvpr2022_Zamir} & {CVPR'22} & 24.09/0.927 & 34.81/0.960 & {31.49}/0.884 & 27.22/0.829 & 20.41/0.806 & 27.60/0.881 \\
AirNet \cite{AirNet_cvpr2022_Li} & {CVPR'22} & 21.04/0.884 & 32.98/0.951 & 30.91/0.882 & 24.35/0.781 & 18.18/0.735 & 25.49/0.846 \\
PromptIR \cite{promptir_nips2023_Vaishnav} & {NeurIPS'23} & 26.54/0.949 & 36.37/0.970 & 31.47/0.886 & 28.71/0.881 & 22.68/0.832 & 29.15/0.904 \\
InstructIR \cite{Instructir_eccv2024_Marcos}& {ECCV'24} & 27.10/0.956 & 36.84/0.973 & 31.40/0.887 & {29.40}/{0.886} & 23.00/0.836 & 29.55/0.907 \\
DiffUIR \cite{DiffUIR_CVPR2024_Zheng}& {CVPR'24} & 29.47/0.965 & 35.98/0.968 & 31.02/0.885 & 27.50/0.845 & 22.32/0.826 & 29.25/0.898 \\
AdaIR \cite{AdaIR_ICLR2025_Cui}&{ICLR'25}  & 30.53/0.978 & 38.02/0.981 & 31.35/0.889 & 28.12/0.858 & 23.00/0.845 & 30.20/0.910 \\
VLU-Net \cite{VLUNet_cvpr2025_Zeng}& {CVPR'25} & 30.84/\underline{0.980} & \underline{38.54}/{0.982} & 31.43/\underline{0.891} & 27.46/0.840 & 22.29/0.833 & 30.11/0.905 \\
MoCE-IR \cite{MoCEir_cvpr2024_Zamfir}& {CVPR'25} & 30.48/0.974 & 38.04/{0.982} & 31.34/0.887 & \textbf{30.05}/\textbf{0.899} & 23.00/{0.852} & 30.58/\underline{0.919} \\
DFPIR \cite{DFPIR_CVPR2025_Tian}&  {CVPR'25} & \underline{31.64}/0.979 & 37.62/0.978 & 31.29/0.889 & 28.82/0.873 & \underline{23.82}/0.843 & {30.64}/0.913 \\
ClearAIR \cite{AAAI2026ClearAIR}& {AAAI'26} &  30.12/0.978 & 38.20/{0.982} & \textbf{31.53}/0.888 & {29.67}/{0.887} & 22.83/0.846 & 30.47/0.916 \\
MIRAGE \cite{ICLR2026MIRAGE}&  {ICLR'26} &  31.45/\underline{0.980} & \textbf{38.92}/\textbf{0.982} & 31.41/\textbf{0.892} & 28.10/0.858 & 23.59/\underline{0.858} & \underline{30.68}/0.914 \\
\midrule
{DAR-Net (Ours)}& - & \textbf{31.67}/\textbf{0.981} & {38.34}/\underline{0.983} & \underline{31.46}/\textbf{0.892} & \underline{29.77}/\underline{0.889} & \textbf{23.86}/\textbf{0.860} & \textbf{31.02}/\textbf{0.921} \\
\bottomrule
\end{tabular}
}
\label{tab:main_5d}
\end{table*}

 \begin{table*}[t]
  \centering
  \vspace{-5pt}
  \caption{Quantitative Comparison on CDD-11 \cite{guo2024CDD_11} Dataset. We reported PSNR/SSIM metrics.}
  \vspace{-10pt}
  \setlength{\tabcolsep}{0.1cm}
    \begin{footnotesize}{
    \resizebox{\textwidth}{!}{
    \begin{tabular}{l|cccc|ccccc|cc|c}
    \toprule
    \multirow{2}{*}{\textbf{Method}} & \multicolumn{4}{c|}{\textbf{Single}} & \multicolumn{5}{c|}{\textbf{Double}} & \multicolumn{2}{c|}{\textbf{Triple}} & \multirow{2}{*}{\textbf{Average}} \\ 
      & \textbf{L} & \textbf{H} & \textbf{R} & \textbf{S} & \textbf{L+H} & \textbf{L+R} & \textbf{L+S} & \textbf{H+R} & \textbf{H+S} & \textbf{L+H+R} & \textbf{L+H+S} &  \\
    \midrule
    AirNet \cite{AirNet_cvpr2022_Li} & 24.83/0.778 & 24.21/0.951 & 26.55/0.891 & 26.79/0.919 & 23.23/0.779 & 22.82/0.710 & 23.29/0.723 & 22.21/0.868 & 23.29/0.901 & 21.80/0.708 & 22.24/0.725 & 23.75/0.814 \\
    PromptIR \cite{promptir_nips2023_Vaishnav} & 26.32/0.805 & 26.10/0.969 & 31.56/0.946 & 31.53/0.960 & 24.49/0.789 & 25.05/0.771 & 24.51/0.761 & 24.54/0.924 & 23.70/0.925 & 23.74/0.752 & 23.33/0.747 & 25.90/0.850 \\
    WeatherDiff \cite{tpami2022WeatherDiff} & 23.58/0.763 & 21.99/0.904 & 24.85/0.885 & 24.80/0.888 &  21.83/0.756 & 22.69/0.730 & 22.12/0.707 & 21.25/0.868 & 21.99/0.868 & 21.23/0.716 & 21.04/0.698 & 22.49/0.799 \\
    WGWS-Net \cite{CVPRLearningWeatherGWS} & 24.39/0.774 & 27.90/0.982 & 33.15/0.964 & 34.43/0.973 & 24.27/0.800 & 25.06/0.772 & 24.60/0.765 & 27.23/0.955 & 27.65/0.960 & 23.90/0.772 & 23.97/0.771 & 26.96/0.863 \\
    OneRestore \cite{guo2024CDD_11} & 26.48/\underline{0.826} & 32.52/\underline{0.990} & 33.40/0.964 & 34.31/0.973 & 25.79/\underline{0.822} & 25.58/0.799 & 25.19/0.789 & \underline{29.99}/0.957 & 30.21/0.964 & 24.78/0.788 & 24.90/0.791 & 28.47/0.878 \\
    AdaIR \cite{AdaIR_ICLR2025_Cui} & 26.88/0.821 & 31.60/0.987 & 33.84/0.962 & 34.65/0.974 & 25.69/0.811 & 25.90/0.793 & 25.69/0.783 & 29.38/0.955 & 28.95/0.961 & 24.82/0.778 & 25.04/0.778 & 28.40/0.873 \\
    MoCE-IR \cite{MoCEir_cvpr2024_Zamfir} & \underline{27.26}/0.824 & \underline{32.66}/\underline{0.990} & \underline{34.31}/\underline{0.970} & \underline{35.91}/\underline{0.980} & \underline{26.24}/0.817 & \underline{26.25}/\underline{0.800} & \underline{26.04}/\underline{0.793} & 29.93/\underline{0.964} & 30.19/\underline{0.970} & \underline{25.41}/\underline{0.789} & \underline{25.39}/\underline{0.790} & \underline{29.05}/\underline{0.881} \\
    \midrule
    DAR-Net (Ours) & \textbf{27.54}/\textbf{0.834} & \textbf{34.21}/\textbf{0.991} & \textbf{34.96}/\textbf{0.972} & \textbf{36.43}/\textbf{0.981} & \textbf{26.74}/\textbf{0.830} & \textbf{26.63}/\textbf{0.812} & \textbf{26.58}/\textbf{0.806} & \textbf{31.25}/\textbf{0.968} & \textbf{31.28}/\textbf{0.971} & \textbf{25.53}/\textbf{0.801} & \textbf{25.83}/\textbf{0.799} & \textbf{29.73}/\textbf{0.888} \\
    \bottomrule
    \end{tabular}
    }}
    \end{footnotesize}    
  \label{tab:cdd11}
\end{table*}

\begin{table*}[t]
  \centering
  \vspace{-5pt}
  \caption{Quantitative Comparison of different methods on WeatherBench \cite{Guan2025WeatherBenchAR} Dataset.}
  \vspace{-10pt}
  \begin{footnotesize}{
  \resizebox{\textwidth}{!}{
    \begin{tabular}{l|c@{\hspace{0.2cm}}c@{\hspace{0.2cm}}c@{\hspace{0.2cm}}c|c@{\hspace{0.2cm}}c@{\hspace{0.2cm}}c@{\hspace{0.2cm}}c|c@{\hspace{0.2cm}}c@{\hspace{0.2cm}}c@{\hspace{0.2cm}}c|c@{\hspace{0.2cm}}c@{\hspace{0.2cm}}c@{\hspace{0.2cm}}c}
    \toprule
    \multirow{2}{*}{\textbf{Method}} & \multicolumn{4}{c|}{\textbf{Dehazing}} & \multicolumn{4}{c|}{\textbf{Deraining}} & \multicolumn{4}{c|}{\textbf{Desnowing}} & \multicolumn{4}{c}{\textbf{Average}} \\
       & \textbf{PSNR$\uparrow$} & \textbf{SSIM$\uparrow$} & \textbf{LPIPS$\downarrow$} & \textbf{FID$\downarrow$} & \textbf{PSNR$\uparrow$} & \textbf{SSIM$\uparrow$} & \textbf{LPIPS$\downarrow$} & \textbf{FID$\downarrow$} & \textbf{PSNR$\uparrow$} & \textbf{SSIM$\uparrow$} & \textbf{LPIPS$\downarrow$} & \textbf{FID$\downarrow$} & \textbf{PSNR$\uparrow$} & \textbf{SSIM$\uparrow$} & \textbf{LPIPS$\downarrow$} & \textbf{FID$\downarrow$} \\
    \midrule
    AirNet \cite{AirNet_cvpr2022_Li} &  19.27  & 0.645  & 0.3829  & 134.09  & 31.56  & 0.912  & 0.2236  & 125.54  & 20.58  & 0.737  & 0.2912  & 138.57  & 23.80  & 0.764  & 0.2992  & 132.73  \\
    TransWeather \cite{CVPR2021TransWeatherTR}  & 18.13  & 0.621  & 0.3970  & 123.21  & 28.59  & 0.880  & 0.2638  & 149.66  & 24.06  & 0.754  & 0.2250  & 102.99  & 23.59  & 0.752  & 0.2953  & 125.29  \\
    PromptIR \cite{promptir_nips2023_Vaishnav} & 19.50  & 0.658  & 0.3751  & 113.55  & 32.51  & 0.915  & 0.1980  & 111.69  & 26.35  & 0.804  & 0.1951  & 84.12  & 26.12  & 0.792  & 0.2561  & 103.12  \\
    WGWS-Net \cite{CVPRLearningWeatherGWS}  & 11.78  & 0.532  & 0.5351  & 152.76  & 34.77  & \underline{0.939}  & \textbf{0.1168}  & \textbf{60.99}  & 19.39  & 0.721  & 0.2481  & 128.56  & 21.98  & 0.731  & 0.3000  & 114.10  \\
    Histoformer \cite{ECCV2024RestoringII}  & 15.82  & 0.597  & 0.4371  & 128.34  & 28.87  & 0.876  & 0.2785  & 152.42  & 23.88  & 0.769  & 0.2252  & 105.82  & 22.86  & 0.747  & 0.3136  & 128.86  \\
    AdaIR \cite{AdaIR_ICLR2025_Cui}  & \underline{21.39}  & 0.680  & \underline{0.3506}  & \underline{110.07}  & 32.81  & 0.918  & 0.1916  & 109.41  & 26.87  & 0.806  & 0.1790  & 73.48  & 27.02  & 0.801  & 0.2404  & 97.65  \\
    DiffUIR \cite{DiffUIR_CVPR2024_Zheng}  & 20.96  & \underline{0.695}  & 0.3550  & 127.54  & \underline{33.78}  & {0.931}  & 0.1720  & 86.96  & \underline{27.87}  & \underline{0.844}  & \underline{0.1619}  & \underline{68.99}  & \underline{27.54}  & \underline{0.823}  &\underline{0.2296} & \underline{94.50}  \\
    \midrule
    DAR-Net (Ours) & \textbf{23.44}  & \textbf{0.732}  & \textbf{0.3257}  & \textbf{108.35}  & \textbf{35.48}  & \textbf{0.941}  & \underline{0.1663}  & \underline{82.65}  & \textbf{29.37}  & \textbf{0.872}  & \textbf{0.1569}  & \textbf{65.28}  & \textbf{29.43}  & \textbf{0.848}  & \textbf{0.2163}  & \textbf{85.43}  \\
    \bottomrule
    \end{tabular}
    }}
    \end{footnotesize}
  \label{tab:weatherbench}
\end{table*}

\subsection{Training Objective}
\label{sec:loss}

DAR-Net is trained with a pixel-wise reconstruction loss and the orthogonality regularization introduced in \S~\ref{sec:spar}. The overall loss is
\begin{equation}
    \mathcal{L}_{\mathrm{total}}
    =
    \mathcal{L}_{\mathrm{rec}}
    +
    \lambda\,\mathcal{L}_{\mathrm{OSR}},
\end{equation}
where $\lambda$ is a balancing coefficient. We adopt the L1 loss between the restored image $I_{\mathrm{out}}$ and the ground-truth image $I_{\mathrm{gt}}$ as the reconstruction loss:
\begin{equation}
    \mathcal{L}_{\mathrm{rec}}
    =
    \frac{1}{N}
    \sum_{i=1}^{N}
    \left|I_{\mathrm{out}}^{(i)}-I_{\mathrm{gt}}^{(i)}\right|,
\end{equation}
where $N$ denotes the total number of image elements. The term $\mathcal{L}_{\mathrm{OSR}}$, defined in Eq.~(\ref{eq:OSR}), regularizes the degradation-aware prompt and the corresponding content feature at the deepest decoder stage by encouraging their spatial response subspaces to be orthogonal.

\section{Experiments}

We evaluate DAR-Net under both three-degradation (3D) and five-degradation (5D) all-in-one restoration settings. Beyond standard evaluation, we further assess its generalization ability on mixed degradations, and real-world images. We compare DAR-Net with representative restoration methods, including Restormer~\cite{Restormer_cvpr2022_Zamir}, AirNet~\cite{AirNet_cvpr2022_Li}, PromptIR~\cite{promptir_nips2023_Vaishnav}, InstructIR~\cite{Instructir_eccv2024_Marcos}, DiffUIR~\cite{DiffUIR_CVPR2024_Zheng}, AdaIR~\cite{AdaIR_ICLR2025_Cui}, VLU-Net~\cite{VLUNet_cvpr2025_Zeng}, MoCE-IR~\cite{MoCEir_cvpr2024_Zamfir}, DFPIR~\cite{DFPIR_CVPR2025_Tian}, ClearAIR~\cite{AAAI2026ClearAIR}, MIRAGE \cite{ICLR2026MIRAGE}, WeatherDiff~\cite{tpami2022WeatherDiff}, WGWS-Net~\cite{CVPRLearningWeatherGWS}, OneRestore~\cite{guo2024CDD_11}, TransWeather \cite{CVPR2021TransWeatherTR} and Histoformer~\cite{ECCV2024RestoringII}. We use PSNR and SSIM~\cite{Zhou2004SSIM} for pixel-wise fidelity evaluation, and LPIPS~\cite{zhang2018LPIPS} and FID~\cite{heusel2017FID} for perceptual quality assessment. Unless otherwise specified, the results of the compared methods are taken from their original papers or from the survey~\cite{jiang2025survey}. The best and second-best results are highlighted in \textbf{bold} and \underline{underlined}, respectively.

\begin{figure*}[!t]
    \captionsetup{aboveskip=3pt, belowskip=-3pt}
    \centering
    \includegraphics[width=1\linewidth]{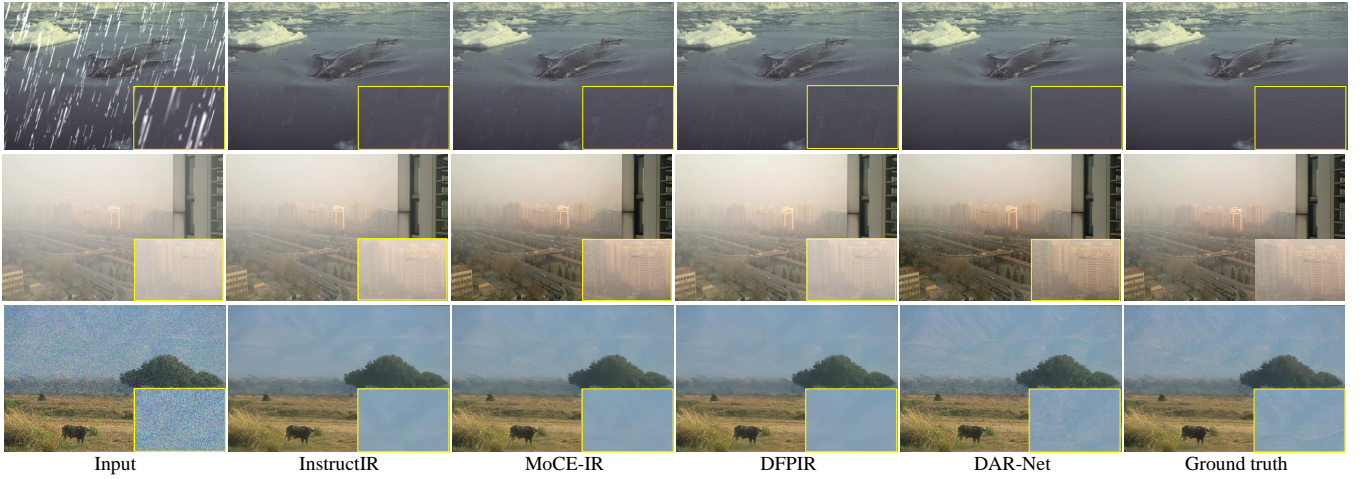}
    \caption{Qualitative comparison of 3D all-in-one restoration results.}
    \Description{A grid of qualitative examples compares image restoration results from multiple methods on the three-degradation all-in-one restoration benchmark.}
    \label{fig:qualitative_5d}
\end{figure*}
\subsection{Experimental Settings}

\noindent\textbf{Datasets.}
For the 3D setting, we train on BSD400~\cite{arbelaez2010BSD400} and WED~\cite{ma2016WED}, and evaluate denoising on BSD68~\cite{martin2001BSD68} with Gaussian noise levels $\sigma \in \{15,25,50\}$. Rain100L~\cite{yang2017Rain100L} and SOTS~\cite{li2018SOTS} are used for deraining and dehazing, respectively. For the 5D setting, we further include GoPro~\cite{nah2017GOPRO} for deblurring and LOL~\cite{wei2018LOL} for low-light enhancement. For mixed-degradation evaluation, we use CDD-11~\cite{guo2024CDD_11}. For real-world evaluation, we adopt the WeatherBench~\cite{Guan2025WeatherBenchAR}.

\noindent\textbf{Implementation details.}
Our model is built on a hierarchical U-shaped Transformer backbone. We use a 4-level encoder-decoder architecture with [4, 6, 6, 8] Transformer blocks from level-1 to level-4. We optimize the network using AdamW with an initial learning rate of $2\times10^{-4}$ and $\beta_1=0.9$, $\beta_2=0.99$. The learning rate is decayed to $1\times10^{-8}$ using cosine annealing with five cycles. The model is trained for 450K and 650K iterations under the 3D and 5D settings, respectively, with a batch size of 32. During training, input images are randomly cropped into $128\times128$ patches and augmented by random flipping and rotation. The number of degradation archetypes $K=16$, the temperature parameter $\tau=0.07$, and the loss weight $\lambda=0.1$. All experiments are implemented in PyTorch and conducted on 2 NVIDIA A800 GPUs.

\subsection{Main Results}
\label{sec:main_results}

\noindent\textbf{Three-Degradation Evaluation.}
As shown in Tab.~\ref{tab:main_3d}, DAR-Net achieves the best overall performance under the 3D setting, with an average PSNR/SSIM of 33.05/0.922. It consistently ranks first on dehazing, deraining, and all three denoising levels, demonstrating strong and balanced restoration performance across different degradation types. Compared with the second-best method, DAR-Net improves the average PSNR by \textbf{0.14 dB}. These results indicate that DAR-Net can more effectively handle degradation-content entanglement in the all-in-one restoration setting, leading to both stronger degradation removal and better content preservation.

\noindent\textbf{Five-Degradation Evaluation.}
DAR-Net achieves the best overall performance under the 5D setting, with an average PSNR/SSIM of 31.02/0.921 (Tab.~\ref{tab:main_5d}). Compared with the second-best method, DAR-Net improves the average PSNR by \textbf{0.34 dB}. Although it is not the best-performing method on every task, DAR-Net achieves the best results on dehazing and low-light enhancement while remaining competitive on deraining, denoising, and deblurring. These results indicate that DAR-Net maintains a strong overall balance across diverse degradation types in the more challenging 5D setting.

\begin{table}[t]
\centering
\begin{minipage}[t]{0.49\linewidth}
\captionsetup{aboveskip=3pt, belowskip=-5pt}
\centering
\caption{Ablation on the key components of DAR-Net.}
\resizebox{\textwidth}{!}{
\begin{tabular}{ccccc}
\toprule \textbf{DAR} & \textbf{SeAR} & \textbf{SpAR} & \textbf{PSNR} & \textbf{SSIM} \\
\midrule
 \ding{55}& \ding{55} & \ding{55} & 29.15 & 0.904 \\
 \ding{51} & \ding{55} & \ding{55} & 29.22 & 0.905 \\
 \ding{51} & \ding{51} & \ding{55} & \underline{30.65} & \underline{0.916} \\
 \ding{51} & \ding{51} & \ding{51} & \textbf{31.02} & \textbf{0.921} \\
\bottomrule
\end{tabular}
}
\label{tab:ablation_overall}
\end{minipage}
\hfill
\begin{minipage}[t]{0.49\linewidth}
\captionsetup{aboveskip=3pt, belowskip=-2pt}
\centering
\caption{Ablation on content feature construction.}
\resizebox{\textwidth}{!}{
\begin{tabular}{l|cc}
\toprule
\textbf{Method} & \textbf{PSNR} & \textbf{SSIM} \\
\midrule
No decomposition & 30.78&0.917 \\
Gated suppression & 30.84&0.918 \\
Residual subtraction & \textbf{31.02}&\textbf{0.921} \\
\bottomrule
\end{tabular}
}
\label{tab:ablation_fc}
\end{minipage}
\end{table}

\noindent\textbf{Mixed-degradation Evaluation.}
As shown in Tab.~\ref{tab:cdd11}, DAR-Net achieves the best results on all CDD-11 \cite{guo2024CDD_11} subsets, covering single, double, and triple degradations. It obtains the highest average PSNR/SSIM of 29.73/0.888, surpassing the second-best method by 0.68 dB in PSNR and 0.007 in SSIM. The consistent gains across increasingly complex degradation combinations verify the effectiveness of DAR-Net for mixed-degradation restoration.

\noindent\textbf{Real-world Evaluation.}
Tab.~\ref{tab:weatherbench} reports the quantitative comparison on the WeatherBench \cite{Guan2025WeatherBenchAR} dataset. DAR-Net achieves the best overall performance, with particularly clear advantages on dehazing and desnowing. On deraining, DAR-Net also attains the best PSNR and SSIM, while remaining competitive in LPIPS and FID. These results demonstrate that DAR-Net generalizes effectively to real-world weather degradations and yields restoration results with improved fidelity and perceptual quality.

\noindent\textbf{Qualitative Results.} Fig.~\ref{fig:qualitative_5d} presents qualitative results under the 3D setting. DAR-Net removes degradations more thoroughly across diverse restoration tasks while better preserving natural image structures. For example, in the deraining case, our result is free of visible rain-streak residue, whereas competing methods still retain noticeable artifacts. In the denoising example with $\sigma=25$, DAR-Net suppresses noise effectively without mistakenly removing the cloud structures in the sky.

\subsection{Ablation Study}
\label{sec:ablation}

\noindent\textbf{Effect of Key Components.}
Tab.~\ref{tab:ablation_overall} reports only the average results under the 5D setting for clarity. The full DAR-Net achieves the best performance, validating the effectiveness of the overall design and the complementarity of its three components. DAR provides a structured degradation prior, while SeAR yields more substantial gains by alleviating channel-wise semantic ambiguity. SpAR further improves the performance, and the combination of all three components leads to the best overall result.

\noindent\textbf{Effect of Content Feature Construction.}
We analyze how to construct the content feature $F_c$ in SpAR while keeping DAR, SeAR, and SpAR enabled. Specifically, we compare three variants: no decomposition ($F_c = F_{in}$), gated suppression ($F_c = F_{in}\odot (1-\sigma(F_{dp}))$), and residual subtraction ($F_c = F_{in} - F_{dp}$). Here, $F_{in}$ denotes the mixed input feature and $F_{dp}$ denotes the degradation-related feature in \S~\ref{sec:spar}. As shown in Tab.~\ref{tab:ablation_fc}, the residual formulation achieves the best performance, suggesting that explicitly subtracting degradation-related information is more effective for isolating content.

\noindent\textbf{Effect of SpAR Placement.}
Applying SpAR at the deepest decoder stage yields the best performance; detailed placement results are provided in the supplementary material.

\begin{figure}[t]
    \centering
    \includegraphics[width=1\linewidth]{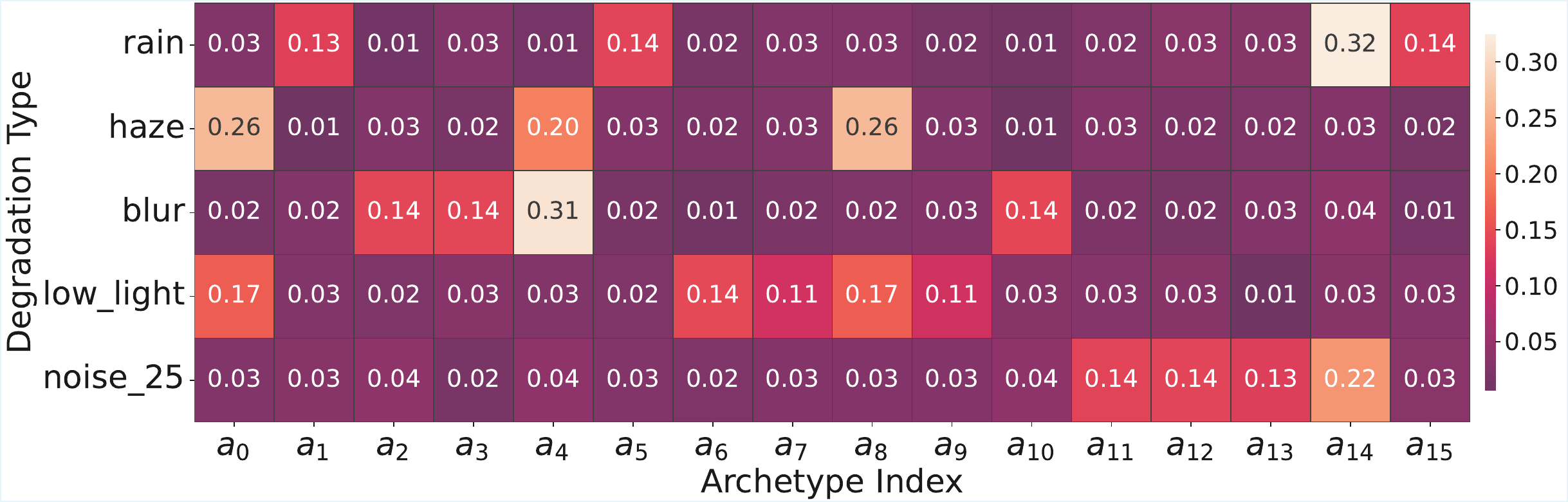}
    \vspace{-20pt}
    \caption{Visualization of the average simplex coefficients learned by DAR for different degradation types. Distinct degradations exhibit different archetype activation patterns, while related degradations still share partial archetypes.}
    \Description{A visualization compares the average simplex coefficients learned by DAR across degradation types. Different degradation types activate distinct archetype patterns, while related degradations share some archetype activations.}
    \vspace{-5pt}
    \label{fig:dar_analysis}
\end{figure}

\begin{figure}[t]
    \centering
    \includegraphics[width=0.7\linewidth]{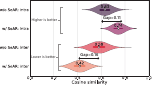}
     \vspace{-10pt}
    \caption{Intra-/inter-class similarity distributions of prompt features before and after SeAR. SeAR increases intra-class similarity while decreasing inter-class similarity.}
    \Description{Similarity distributions of prompt features before and after SeAR show that SeAR shifts intra-class similarity upward and inter-class similarity downward.}
    \vspace{-10pt}
    \label{fig:intra_inter}
\end{figure}

\subsection{Analysis}

\noindent\textbf{Analysis of DAR.}
Fig.~\ref{fig:dar_analysis} shows that different degradations activate distinct archetype combinations, while related degradations share partial archetypes, indicating structured yet transferable degradation representations. Analysis of the archetype number $K$ is provided in the supplementary material.

\noindent\textbf{Analysis of SeAR.}
As shown in Fig.~\ref{fig:intra_inter}, SeAR increases intra-class prompt similarity from 0.63 to 0.74 and decreases inter-class similarity from 0.58 to 0.42, demonstrating improved degradation discrimination.

\begin{figure}[!t]
    \captionsetup{aboveskip=3pt, belowskip=-3pt}
    \centering
    \includegraphics[width=1\linewidth]{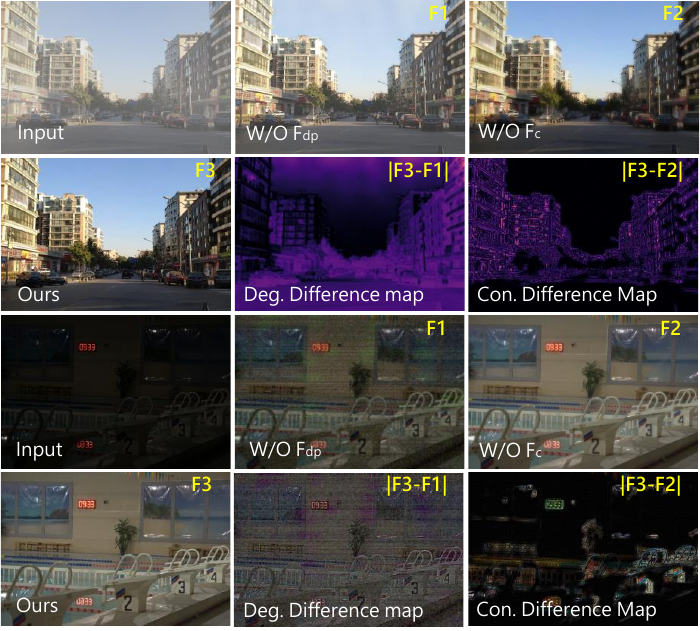}
    \caption{Analysis of SpAR. F1, F2, and F3 denote the results without the degradation-related feature, without the content-related feature, and with the full model, respectively. The difference maps indicate distinct roles of the two features in degradation removal and content preservation.}
    \Description{SpAR analysis compares F1 without the degradation-related feature, F2 without the content-related feature, and F3 from the full model. Difference maps show that the two features play distinct roles in removing degradation and preserving content.}
    \label{fig:spar_analysis}
\end{figure}

\noindent\textbf{Analysis of SpAR.}
As shown in Fig.~\ref{fig:spar_analysis}, removing $F_{dp}$ leaves residual degradations, whereas removing $F_c$ damages structural content, confirming their complementary roles in degradation removal and content preservation.

\noindent\textbf{Model Complexity and Efficiency.}
As shown in Tab.~\ref{tab:para}, DAR-Net has 35.5M parameters and 771G FLOPs, which are comparable to existing methods. Despite slightly higher complexity than some lightweight baselines, DAR-Net still achieves competitive inference efficiency, with lower latency than PromptIR, AdaIR, and DFPIR. Compared with MoCE-IR, DAR-Net incurs moderate additional overhead while providing stronger restoration performance, demonstrating a favorable efficiency-performance trade-off.

\begin{table}[!t]
		\centering
		\caption{Comparison of model complexity and inference efficiency. FLOPs and latency are measured on an input image of size $720\times480$ on a single NVIDIA A800 GPU.}
	\vspace{-10pt}
	\resizebox{\linewidth}{!}{
        \begin{tabular}{l|ccccc}
			\toprule
			\textbf{Method}  & \textbf{PromptIR} & \textbf{AdaIR} & \textbf{MoCE-IR} & \textbf{DFPIR} & \textbf{DAR-Net} \\
			\midrule
			Params. &34.1M&28.8M & 25.4M & 31M + 63M & 35.5M\\
			FLOPs   &752G &786G  &493G      &  885G  &771G\\
			Latency &187ms&239ms &161ms     &193ms   &165ms\\
         CPU Memory &4454M&4336M &4389M     &4369M   &4200M\\
         GPU Memory &3324M&3117M &1445M     &3459M   &3162M\\
			\bottomrule
    \end{tabular}
    \label{tab:para}
	}
\end{table}

\section{Conclusion}

In this paper, we presented DAR-Net, a dual-ambiguity rectification network for all-in-one image restoration. We identify that existing unified restoration methods often suffer from semantic ambiguity in channel-wise representations and spatial ambiguity in spatial responses. To address this, we introduced DAR to learn structured degradation states, SeAR to improve channel-wise degradation discrimination, and SpAR to reduce spatial entanglement between degradation and content. Extensive experiments demonstrate that DAR-Net achieves strong and balanced restoration performance across diverse degradations. These results suggest that explicitly modeling what should be removed and what should be preserved is an effective direction for unified image restoration.

\begin{acks}
This research was partially supported by the National Natural Science Foundation of China (NSFC) (62306064) and the Sichuan Science and Technology Program (granted No. 2024ZDZX0011, No. 2026NSFSC1482 and No. 2025ZHCG0002).
\end{acks}

\bibliographystyle{ACM-Reference-Format}
\bibliography{sample-base}


\end{document}